\pdfoutput=1

\documentclass[11pt]{article}

\usepackage[final]{acl}

\usepackage{times}
\usepackage{latexsym}

\usepackage[T1]{fontenc}

\usepackage[utf8]{inputenc}

\usepackage{microtype}

\usepackage{inconsolata}

\usepackage{graphicx}
\usepackage{amsmath} 
\usepackage{amssymb}  
\usepackage{hyperref} 

\usepackage[table]{xcolor}
\usepackage{amsmath}
\usepackage{subcaption}
\usepackage{fvextra}
\usepackage[title]{appendix}
\usepackage{multirow}
\usepackage{multicol}
\usepackage{booktabs}
\usepackage{makecell}

\usepackage{enumitem}
\usepackage{ltxtable}
\usepackage{listings}
\usepackage{comment}
\usepackage{fancyvrb}
\usepackage{caption}
\usepackage{cprotect}
\usepackage{fontawesome5}
\usepackage{bm}
\usepackage{array}

\usepackage{algorithm}
\usepackage{algorithmic}
\usepackage{xcolor}

\usepackage{booktabs}    

\DefineVerbatimEnvironment{Verbatim}{Verbatim}{breaklines=true}

\usepackage{inconsolata}
\newif\ifshowcomments
\newif\ifarxiv
\arxivtrue
\showcommentstrue 
\ifshowcomments
    \newcommand{\jz}[1]{{\color{blue}[JZ: #1]}}
\else
    \newcommand{\jz}[1]{}
   \ifarxiv
    \newcommand{\app}[1]{#1}
    \else
    \newcommand{\app}[1]{}
    \fi
\fi

\usepackage{xcolor}
\definecolor{darkblue}{RGB}{0, 0, 139}      
\definecolor{darkgreen}{RGB}{0, 100, 0}     

\title{From Reasoning to Answer: Empirical, Attention-Based and Mechanistic Insights into Distilled DeepSeek R1 Models}

\author{Jue Zhang\textsuperscript{1}\thanks{Correspondence to: juezhang@microsoft.com.}, Qingwei Lin\textsuperscript{1}, Saravan Rajmohan\textsuperscript{1}, Dongmei Zhang\textsuperscript{1}\\ 
\textsuperscript{1} Microsoft}

\begin{document}
\maketitle
\begin{abstract}
Large Reasoning Models (LRMs) generate explicit reasoning traces alongside final answers, yet the extent to which these traces influence answer generation remains unclear. In this work, we conduct a three-stage investigation into the interplay between reasoning and answer generation in three distilled DeepSeek R1 models. First, through empirical evaluation, we demonstrate that including explicit reasoning consistently improves answer quality across diverse domains. Second, attention analysis reveals that answer tokens attend substantially to reasoning tokens, with certain mid-layer \emph{Reasoning-Focus Heads} (RFHs) closely tracking the reasoning trajectory, including self-reflective cues. Third, we apply mechanistic interventions using activation patching to assess the dependence of answer tokens on reasoning activations. Our results show that perturbations to key reasoning tokens can reliably alter the final answers, confirming a directional and functional flow of information from reasoning to answer. These findings deepen our understanding of how LRMs leverage reasoning tokens for answer generation, highlighting the functional role of intermediate reasoning in shaping model outputs. Our data and code are publicly available at \href{https://aka.ms/R2A-code}{this URL}.
\end{abstract}

\section{Introduction}

Recent progress in Large Language Models (LLMs) has led to the development of Large Reasoning Models (LRMs), such as OpenAI’s o1~\cite{openai2024openaio1card} and DeepSeek’s R1~\cite{deepseekai2025deepseekr1incentivizingreasoningcapability}, which generate explicit intermediate reasoning traces before producing final answers. This approach, rooted in early Chain-of-Thought (CoT)~\cite{NEURIPS2022_9d560961} prompting techniques, exemplifies a form of test-time scaling, which improves model performance by allocating additional computation during inference.

LRMs typically output two distinct segments: a \textit{Reasoning} segment consisting of reasoning tokens often enclosed within \texttt{<think>} and \texttt{</think>}, and an \textit{Answer} segment providing the final, self-contained response to the query. This separation gives rise to a fundamental question: \textit{Do LRMs actually leverage the reasoning tokens to generate answers?} It is conceivable that the reasoning traces serve merely as post-hoc justifications, rather than functioning as essential components in answer generation~\cite{lanham2023measuringfaithfulnesschainofthoughtreasoning}. This question is closely tied to active research areas including reasoning effectiveness~\cite{ma2025reasoningmodelseffectivethinking}, reasoning faithfulness~\cite{chen2025reasoningmodelsdontsay}, and model behavior monitoring~\cite{baker2025monitoring}. Despite growing efforts in enhancing the overall capabilities of LRMs, the interplay between reasoning and answer segments remains poorly understood.

\begin{table*}[ht]
  \centering
  \small  
  \renewcommand{\arraystretch}{1.2}
  \begin{tabular}{>{\centering\arraybackslash}m{0.3\textwidth} | >{\centering\arraybackslash}m{0.65\textwidth}}
    \toprule
    \textbf{Investigation Directions} & \textbf{Key Observations and Insights} \\
    \midrule
    \textbf{Empirical Study}: Does Reasoning Improve Answer Quality? \newline \textit{(Section~\ref{sec:empirical_evaluation})} &
    \begin{minipage}[c]{\linewidth}
      \begin{itemize}[leftmargin=*, itemsep=0pt, topsep=0pt, parsep=0pt, partopsep=0pt]
        \item Explicit reasoning improves answer quality across diverse domains.
        \item Gains are more pronounced in distilled R1 models than in the full R1 model.
      \end{itemize}
    \end{minipage} \\
    \midrule
    \textbf{Attention Analysis}: How Does Answer Attend to Reasoning? \newline \textit{(Section~\ref{sec:attention_analysis})} &
    \begin{minipage}[c]{\linewidth}
      \begin{itemize}[leftmargin=*, itemsep=0pt, topsep=0pt, parsep=0pt, partopsep=0pt]
        \item Answer tokens strongly attend to reasoning tokens.
        \item Certain mid-layer attention heads closely track reasoning progression.
        \item Attention paths often terminate at the positions of reflection-related tokens.
        \item Attention sink effects persist after reasoning-related model distillation.
        \item \texttt{<think>} and \texttt{</think>} tokens function more likely as structural markers.
        \item Potential application of reasoning-focus heads in reasoning failure debugging.
      \end{itemize}
    \end{minipage} \\
    \midrule
    \textbf{Mechanistic Intervention}: Can Small Reasoning Changes Shift the Answer? \newline \textit{(Section~\ref{sec:mechanistic_intervention})} &
    \begin{minipage}[c]{\linewidth}
      \begin{itemize}[leftmargin=*, itemsep=0pt, topsep=0pt, parsep=0pt, partopsep=0pt]
        \item Modifying the activations of key reasoning tokens can flip the answer.
        \item Evidence suggests reasoning-to-answer information flow in mid-model layers.
      \end{itemize}
    \end{minipage} \\
    \bottomrule
  \end{tabular}
  \caption{Summary of findings across empirical studies, attention analysis, and mechanistic interventions.}
  \label{tab:insight_summary}
\end{table*}

To address this gap, we conduct a three-stage progressive analysis to examine how reasoning contributes to answer generation in Large Reasoning Models, with key insights summarized in Table~\ref{tab:insight_summary}. We focus on three distilled DeepSeek R1 models, i.e., R1-Llama-8B, R1-Qwen-7B, and R1-Qwen-1.5B~\cite{deepseekai2025deepseekr1incentivizingreasoningcapability}, chosen for their accessible reasoning traces and moderate model sizes. We begin with an empirical evaluation comparing model performance with and without reasoning traces. Results show that including reasoning generally improves answer quality across diverse domains, extending prior work that focused primarily on the math and code domains~\cite{ma2025reasoningmodelseffectivethinking}.

While these results indicate that explicit reasoning \textit{can} enhance answer quality, it remains unclear \textit{how} answer tokens incorporate information from reasoning tokens. Since attention mechanisms govern information flow in transformer-based models, we next analyze the attention patterns between the reasoning and answer segments. We find that answer tokens attend substantially to reasoning tokens, and that certain \emph{Reasoning-Focus Heads} (RFHs) consistently track the reasoning process, including self-reflective steps. We further present a case study showing that RFHs can make the source of reasoning errors far more interpretable than head-averaged attention, showing their potential application in debugging failures in reasoning traces.

Finally, since strong attention alone does not guarantee functional dependence, we perform \textit{mechanistic interventions} by perturbing reasoning traces in a controlled setting. These experiments reveal that small modifications to reasoning can indeed flip the answer output.

Our contributions are summarized as follows:
\begin{itemize}[leftmargin=5pt]
\item We broaden the empirical investigation of the influence of reasoning on answer quality in LRMs to a wider range of domains beyond the commonly studied mathematics and code.
\item We analyze attention patterns between reasoning and answer in LRMs, uncovering novel findings such as specific attention heads tracking the progression of reasoning during answer generation.
\item We augment the attention analysis with a detailed mechanistic intervention, demonstrating that perturbations to the activations of reasoning tokens can directly influence the generated answers.
\end{itemize}

\section{Related Work}

\noindent \textbf{Performance Impact of Reasoning in Large Reasoning Models.} The advent of LRMs has spurred interest in analyzing the relationship between reasoning and model performance empirically~\cite{muennighoff2025s1simpletesttimescaling, marjanović2025deepseekr1thoughtologyletsthink, ma2025reasoningmodelseffectivethinking, ballon2025relationshipreasoningperformancelarge, su2025underthinkingoverthinkingempiricalstudy, jahin2025unveilingmathematicalreasoningdeepseek}. While most studies examine the correlation between reasoning length and answer quality, we do not constrain reasoning length, focusing instead on overall response quality. The study most related to ours is~\cite{ma2025reasoningmodelseffectivethinking}, which also compares models with and without reasoning. However, their emphasis still lies in efficiency, with evaluations confined to math and code. Our study extends this comparison to broader open-domains, demonstrating the general improvements in answer quality brought by explicit reasoning.

\noindent \textbf{Measuring Faithfulness in Model Reasoning.} A model's reasoning is considered faithful if it accurately reflects the model’s internal decision-making process~\cite{jacovi-goldberg-2020-towards}. A commonly used method to assess faithfulness is to alter the reasoning tokens and observe how these changes affect the model’s output. Although some argue that this approach primarily evaluates self-consistency rather than true faithfulness~\cite{parcalabescu2024measuringfaithfulnessselfconsistencynatural}, several recent studies~\cite{lanham2023measuringfaithfulnesschainofthoughtreasoning, atanasova-etal-2023-faithfulness, NEURIPS2023_ed3fea90} have employed it to assess the faithfulness of chain-of-thought reasoning in non-LRMs. For example, four manipulation strategies (i.e., early answering, introducing errors, paraphrasing, and adding filler tokens) were used to examine the impact on model predictions~\cite{lanham2023measuringfaithfulnesschainofthoughtreasoning}. More recent work has extended such analyses to LRMs~\cite{baker2025monitoring, arcuschin2025chainofthoughtreasoningwildfaithful, chen2025reasoningmodelsdontsay, chua2025deepseekr1reasoningmodels, marjanović2025deepseekr1thoughtologyletsthink}. For instance, \cite{chen2025reasoningmodelsdontsay} used paired prompts, with and without a hint, to test whether the model explicitly acknowledges using the hint.

In our empirical evaluation, we also manipulate the reasoning content by comparing the with and without reasoning settings, a method loosely related to the early answering strategy in~\cite{lanham2023measuringfaithfulnesschainofthoughtreasoning}. However, our objective differs: instead of measuring the answer flip rate to infer faithfulness, we focus on changes in overall accuracy to assess the functional contribution of reasoning.


\noindent \textbf{Mechanistic Interpretability of Model Reasoning.} Mechanistic interpretability aims to understand how neural networks operate by identifying the specific internal computations and structures responsible for their behavior~\cite{olah2020zoom, meng2022locating, NEURIPS2021_4f5c422f, geva2021transformerfeedforwardlayerskeyvalue, bereska_mechanistic_2024, mueller2025questrightmediatorsurveying}. Understanding model reasoning has been a central focus within this field. Prior to the emergence of LRMs, several studies employed various mechanistic interpretability techniques to analyze chain-of-thought reasoning~\cite{NEURIPS2024_c50f8180, hou-etal-2023-towards, wang2024how, dutta2024thinkstepbystepmechanisticunderstanding, tan-2023-causal, li2024focusquestioninterpretingmitigating}. For instance, attention analysis has been used to reconstruct the model’s reasoning tree in multi-step reasoning tasks~\cite{hou-etal-2023-towards}. In our work, we similarly leverage attention pattern analysis, but specifically to investigate how final answers attend to intermediate reasoning steps.

Following the development of LRMs, a number of mechanistic interpretability studies have explored reasoning-related features specific to these models~\cite{galichin2025icoveredbaseshere, baek2025understandingdistilledreasoningmodels, GoodfireResearch, anthropic2025circuit, venhoff2025understanding}. For instance, sparse features in the MLP layers have been identified and manipulated to influence reasoning behaviors~\cite{galichin2025icoveredbaseshere}. In contrast to these approaches, our work centers on tracing the flow of information from reasoning steps to the final answer, emphasizing the study on the attention modules of LRMs.

\section{Empirical Evaluation: Does Reasoning Improve Answer Quality?}
\label{sec:empirical_evaluation}

\begin{figure}[t]
  \centering
  \begin{subfigure}{\linewidth}
  \centering
    \includegraphics[width=0.95\linewidth]{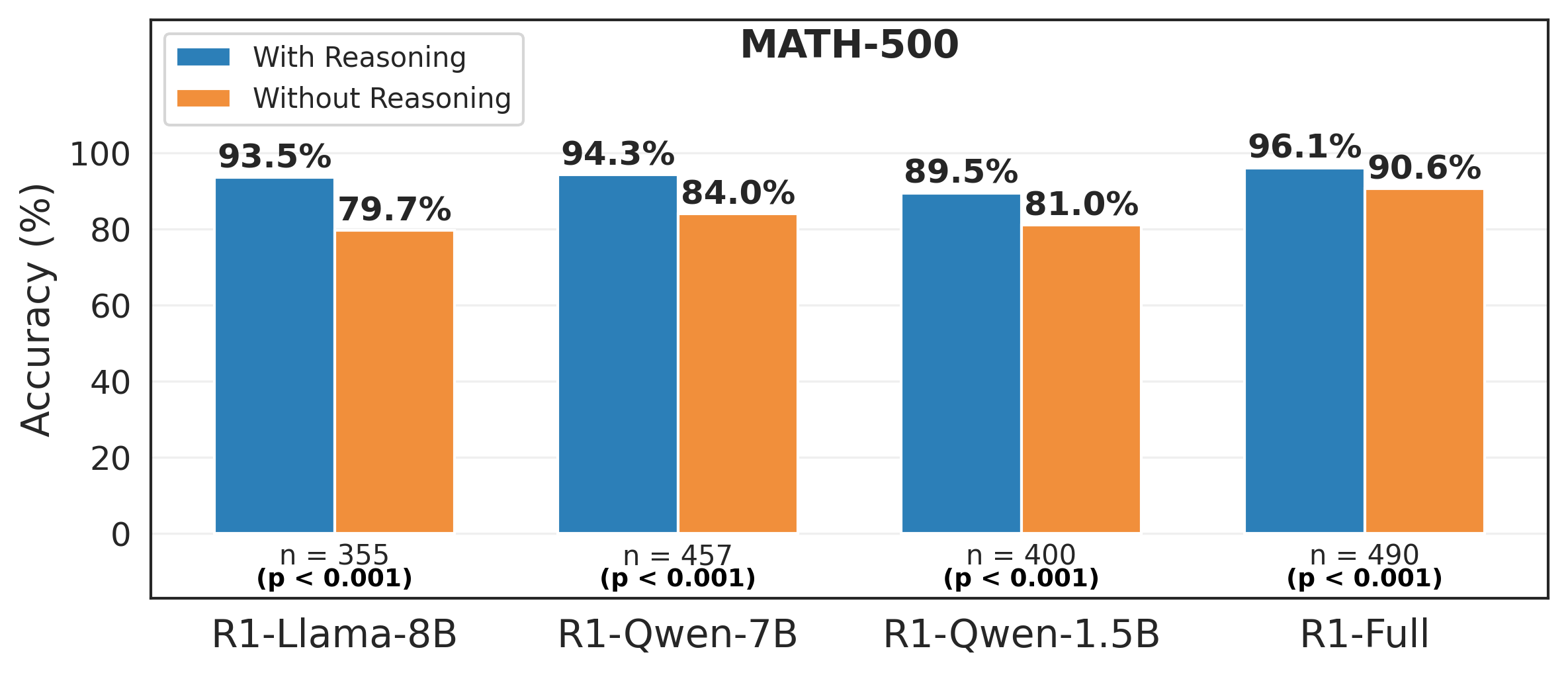}
    \caption{MATH-500: Answer accuracy across R1 model variants.}
  \end{subfigure}
  \hfill
  \vspace{0.5pt}
  \begin{subfigure}{\linewidth}
  \centering
    \includegraphics[width=0.95\linewidth]{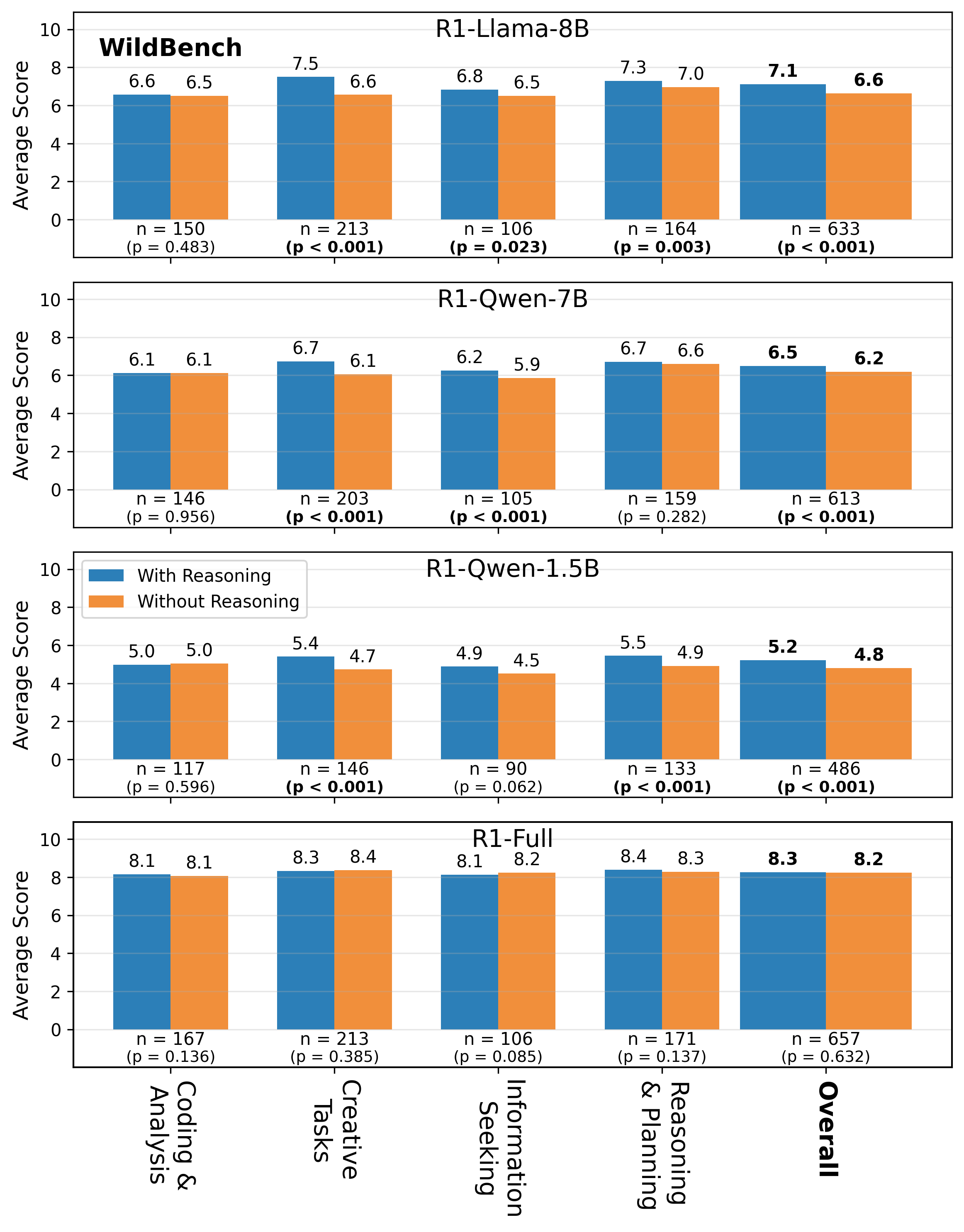}
    \caption{WildBench: Average scores by task domain and model.}
  \end{subfigure}
  \caption{Answer quality comparison for DeepSeek R1 models under reasoning and non-reasoning settings, evaluated on the MATH-500 and WildBench datasets. All statistics are computed over $n = a$ samples, and p-values are obtained using a paired t-test. Statistically significant results with $p < 0.05$ are shown in bold.}
  \label{fig:quality_impact}
\end{figure}

We begin by empirically evaluating whether and how reasoning traces influence answer quality, treating the models as black boxes. This serves as an initial assessment of the impact of reasoning on final outputs. We adopt two datasets: MATH-500~\cite{NEURIPS_DATASETS_AND_BENCHMARKS2021_be83ab3e, lightman2024lets} for mathematics and WildBench~\cite{Lin2024WildBenchBL} for real-world queries that cover diverse domains. Evaluations are conducted under two settings: \textbf{with and without reasoning traces}. For the latter, we adopt the suppression method from~\cite{ma2025reasoningmodelseffectivethinking}, using \textit{``<think>\textbackslash nOkay, I think I have finished thinking.\textbackslash n</think>''} to bypass reasoning.\footnote{We also experimented with \textit{``<think>\textbackslash n\textbackslash n</think>''}, but found it less effective in suppressing reasoning.} 

Following the recommendations in~\cite{deepseekai2025deepseekr1incentivizingreasoningcapability}, we append the instruction \textit{``Please reason step by step, and put your final answer within \textbackslash boxed\{\}.''} to all math-related queries in the MATH-500 dataset and set the sampling temperature to 0.6 for all R1 model variants. We employ zero-shot prompting for both the MATH-500 and WildBench datasets and adopt the evaluation metrics defined in the original benchmarks. Additional experimental details are provided in Appendix~\ref{app:quality_impact}.

Figure~\ref{fig:quality_impact} presents evaluation results for three distilled DeepSeek R1 models, alongside the full R1 model (R1-Full) for reference. Several key observations emerge. First, across nearly all models and domains, incorporating reasoning traces leads to improved answer quality, especially on the MATH-500 dataset, where distilled models show $\sim 10\%$ increase in accuracy, compared to $\sim 5\%$ for the full model. Second, in the general (non-math) domains, such as the ``Overall'' score in Figure~\ref{fig:quality_impact}(b) representing the average performance across all task categories in WildBench, the performance gains from reasoning are again more substantial for the distilled models, whereas R1-Full shows only marginal improvement. This suggests that the full R1 model may already possess sufficient general knowledge, making explicit reasoning during inference less impactful in general domains. 

These results, in conjunction with similar trends observed in other math and code-related datasets~\cite{ma2025reasoningmodelseffectivethinking}, indicate that reasoning tokens contribute to answer quality across diverse tasks, with a more pronounced effect in distilled R1 models compared to the full model. 

\section{Attention Analysis: How Does Answer Attend to Reasoning?}
\label{sec:attention_analysis}

\begin{figure}[t]
  \centering
    \includegraphics[width=\linewidth]{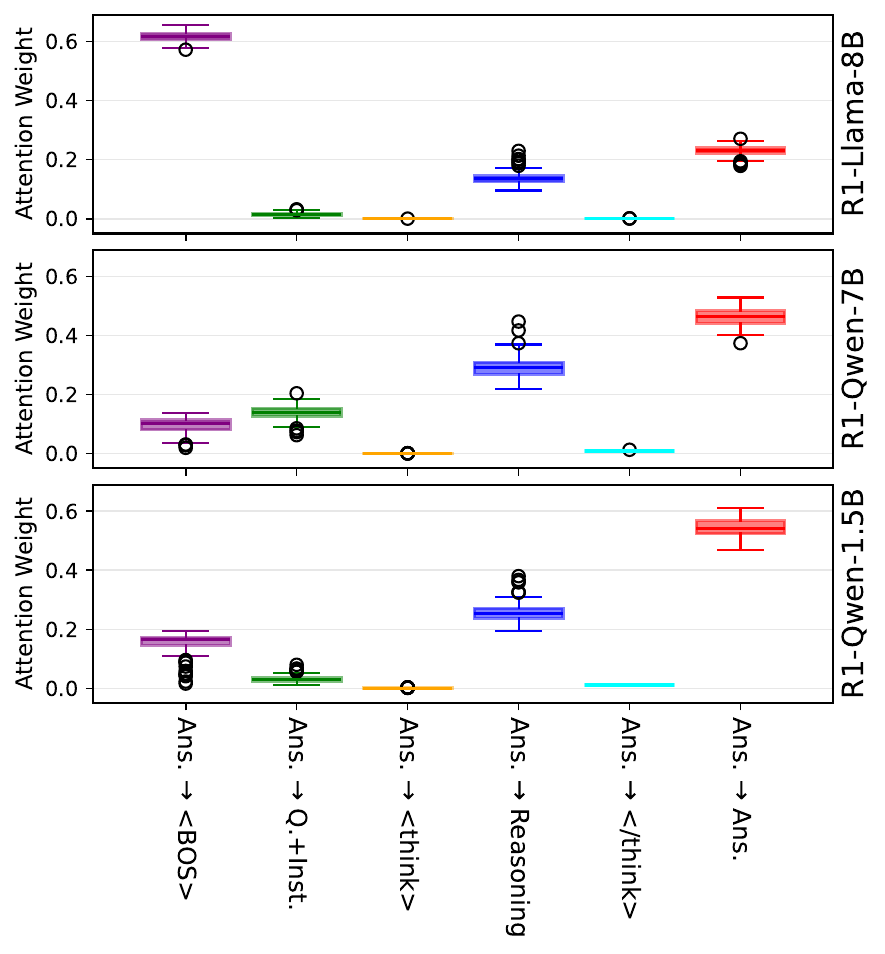}   
  \caption{Decomposition of average attention weights from answer tokens to different prompt segments across three models (R1-Llama-8B, R1-Qwen-7B, R1-Qwen-1.5B) using the MATH-500 dataset.}
  \label{fig:answer_attn_decompose_MATH500}
\end{figure}

\begin{figure*}[t]
    \centering
    \includegraphics[width=\linewidth]{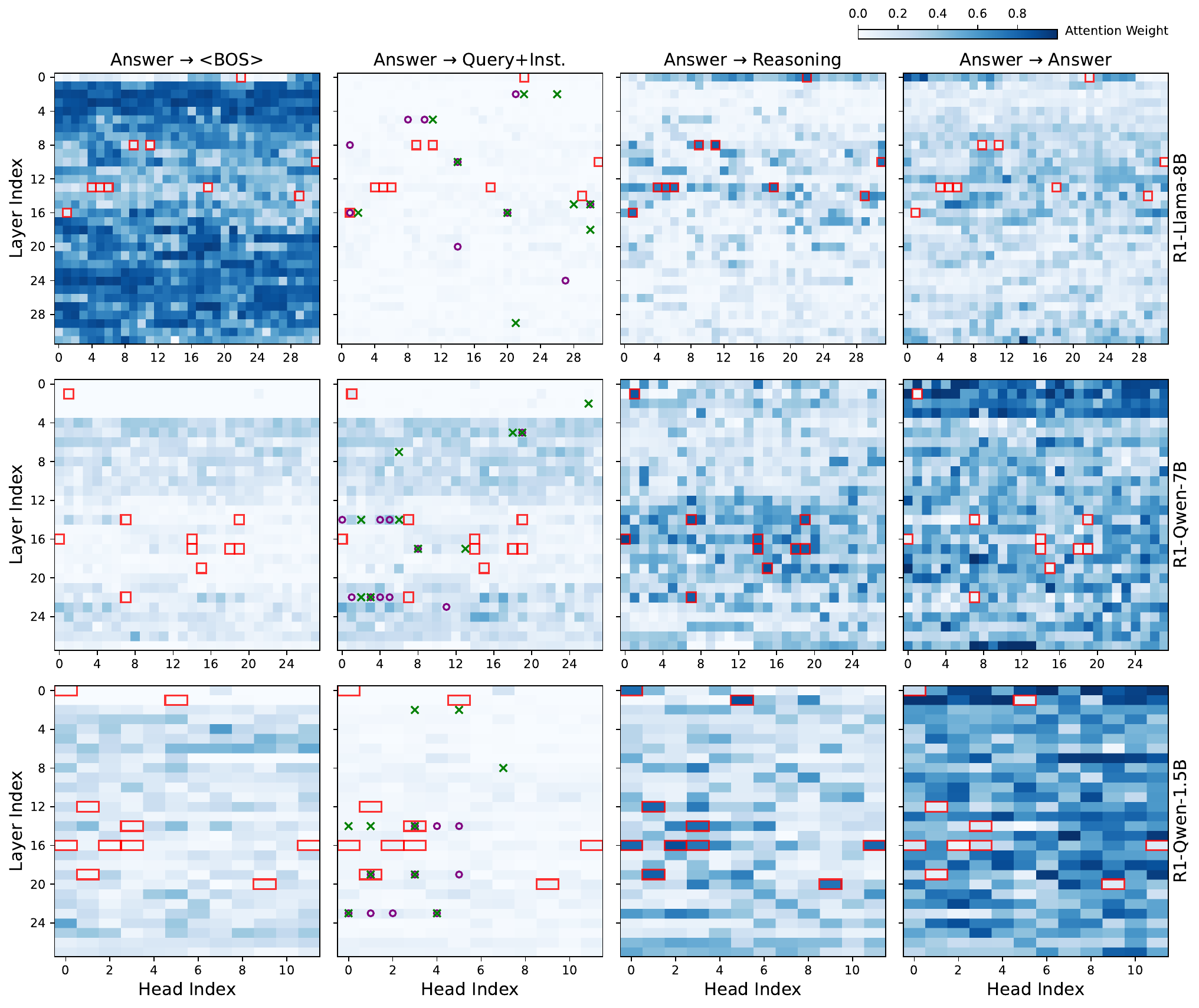}
  \caption{Decomposition of answer attention to different prompt segments across layers and heads for three models (R1-Llama-8B, R1-Qwen-7B, and R1-Qwen-1.5B) using the MATH-500 dataset. Heatmaps show the attention weights from the \textit{Answer} segment to four prompt segments: \texttt{<BOS>}, \textit{Query+Instruction}, \textit{Reasoning}, and \textit{Answer}. Red boxes highlight the top 10 attention heads with the highest weights for \textit{Answer} $\rightarrow$ \textit{Reasoning}. Additionally, in the second column the top 10 retrieval and induction heads are annotated with ``{\color{purple}{$\boldsymbol{\circ}$}}'' and ``{\color{darkgreen}{$\boldsymbol{\times}$}}'', respectively.}
  \label{fig:attn_map_MATH500}
\end{figure*}

Having established that reasoning traces enhance answer quality, we now turn to a more detailed analysis of the model's internal behavior. Since our goal is to understand how information flows from the reasoning to the final answer, we focus on the model's attention mechanisms. Specifically, we analyze how answer tokens attend to different parts of the prompt. Here, the term \textit{“prompt”} refers to the input query and instructions plus the model’s complete response, since all are required to generate the attention patterns. Concretely, we treat the \textit{“prompt”} as a token sequence composed of six segments: \texttt{<BOS>}, \textit{Query+Instruction (QI)}, \texttt{<think>}, \textit{Reasoning}, \texttt{</think>}, and \textit{Answer}, where \texttt{<BOS>} denotes the beginning-of-sentence token.

Our attention analysis builds on the answer quality traces introduced in the previous section, using 100 samples per dataset and per R1-distilled model. We first provide an overall analysis at the prompt segment level, followed by a more detailed analysis across model layers and attention heads. We also include a representative failure case study to illustrate how attention patterns may reveal the source of an incorrect answer. Due to space limitations, we present results for MATH-500 in the main text and defer the qualitatively similar results for WildBench to Appendix~\ref{app:attn_analysis_wildbench}.

\subsection{By Prompt Segment}

Figure~\ref{fig:answer_attn_decompose_MATH500} presents the decomposition of attention weights from answer tokens to various prompt segments. These weights are computed by first averaging over tokens within the \textit{Answer} segment and then aggregating attention towards each destination segment. The results are further averaged across all model layers and attention heads. Based on this aggregated analysis, we observe the following:

\begin{itemize}[leftmargin=5pt]
\item All three R1-distilled models allocate substantial attention from the \textit{Answer} segment to the \textit{Reasoning} segment (blue box-plots), although this cross-segment interaction is weaker than the intra-segment attention within \textit{Answer} (red box-plots).
\item The known attention sink phenomenon~\cite{xiao2024efficient, gu2025attentionsinkemergeslanguage, barbero2025llmsattendtoken} appears in all three R1-distilled models, with noticeable attention allocated to the \texttt{<BOS>} token (purple box-plots). This indicates that the attention sink mechanism persists after reasoning distillation via supervised fine-tuning.
\item Attentions to the \texttt{<think>} and \texttt{</think>} tokens (orange and cyan box-plots) are minimal, suggesting that their primary role is to demarcate different prompt segments rather than to store or summarize preceding information.
\item Lastly, attention to the \textit{Query+Instruction} segment (green box-plots) is relatively low compared to the \textit{Reasoning} and \textit{Answer} segments. This may be attributed to the greater token distance between the \textit{QI} and \textit{Answer} segments, as well as the shorter average query length in the MATH-500 dataset.
\end{itemize}

Overall, this segment-level attention analysis reveals that reasoning tokens receive substantial attention from answer tokens, suggesting their significant role in the answer generation process.

\subsection{Across Model Layers and Attention Heads}

\begin{figure}[t]
    \centering
    \includegraphics[width=\linewidth]{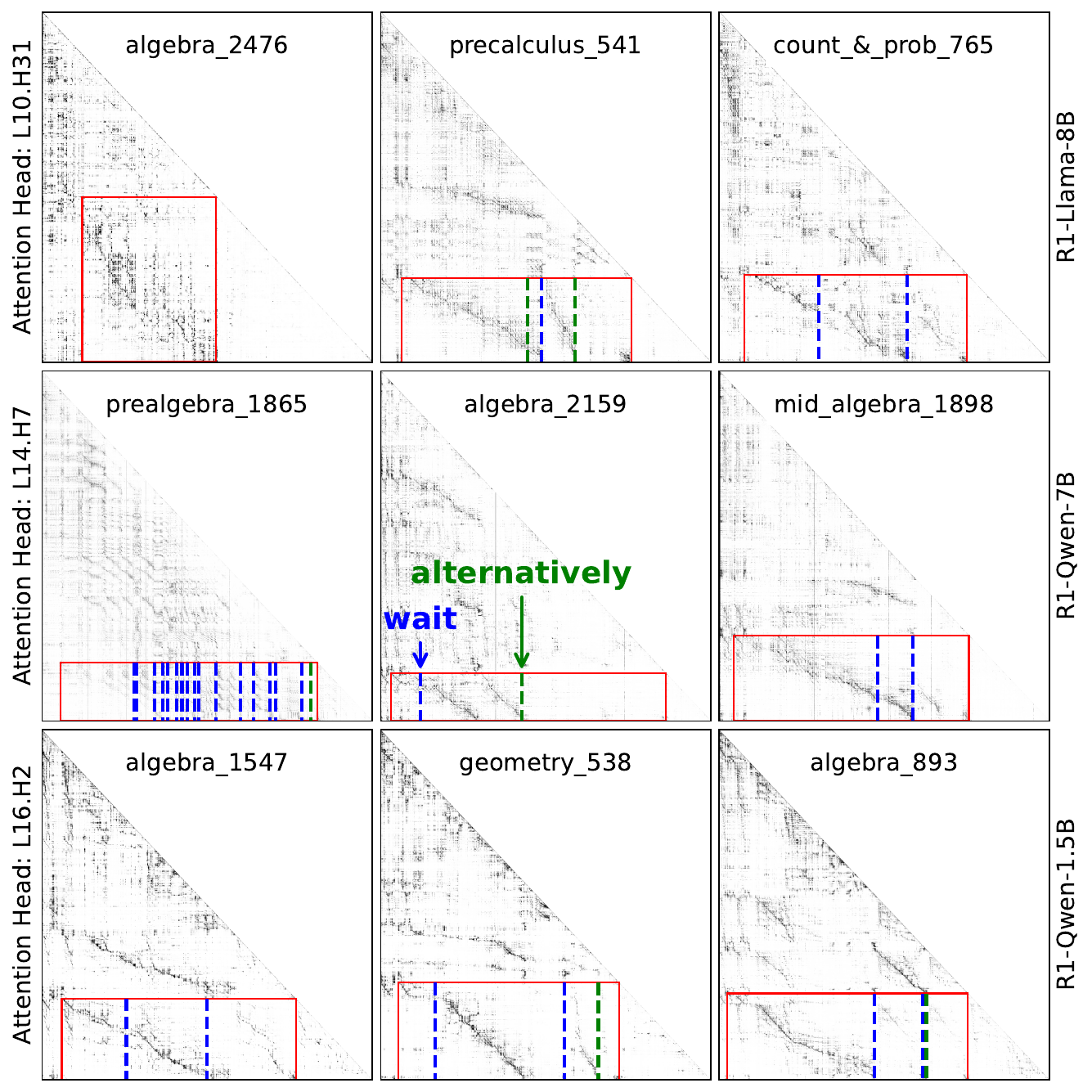}
    \caption{Attention patterns from selected top attention heads of the three distilled R1 models on sample cases from the MATH-500 dataset. The horizontal and vertical axes represent key and query token indices, respectively. Labels on the left y-axis denote the selected reasoning-focused heads (e.g., \textit{L10.H31''} refers to head 31 in Layer 10). The \textit{Answer} $\rightarrow$ \textit{Reasoning} region is highlighted with red boxes. Vertical lines indicate the positions of the tokens ``\textcolor{darkblue}{wait}'' and ``\textcolor{darkgreen}{alternatively}''.}
  \label{fig:attn_pattern_case_study_MATH500_withR}
\end{figure}

Figure~\ref{fig:attn_map_MATH500} shows the decomposition of attention weights from the \textit{Answer} segment to other prompt segments across model layers and attention heads. The top 10 attention heads with the highest attention weights for the \textit{Answer} $\rightarrow$ \textit{Reasoning} are highlighted with red boxes. Notably, these heads are primarily concentrated in the middle layers of the models, i.e., \textbf{Layers 8–16} for R1-Llama-8B, \textbf{Layers 14–22} for R1-Qwen-7B, and \textbf{Layers 12–20} for R1-Qwen-1.5B. This pattern aligns with the prevailing understanding that middle layers in LLMs are primarily responsible for comprehension and reasoning by processing information from lower layers while shaping it for decision-making and generation in later layers~\cite{ju-etal-2024-large}. We therefore refer to these heads as \textbf{Reasoning-Focus Heads (RFHs)}, a term whose relevance will become more apparent in subsequent case studies.

While some attention heads in early layers (e.g., Head 22 at Layer 0 for R1-Llama-8B) also receive large attention of \textit{Answer} $\rightarrow$ \textit{Reasoning}, closer inspection reveals that these typically exhibit a near-uniform focus on preceding tokens such as punctuation and prepositions. More details are provided in the Appendix~\ref{app:early_layer_attn_head}. This behavior can result in larger aggregate attention to longer segments like \textit{Reasoning}, due to a length bias rather than genuine reasoning focus. In contrast, the middle-layer heads are less affected by this length-based artifact.

For comparison, the second column of Figure~\ref{fig:attn_map_MATH500} presents the top 10 induction heads~\cite{olsson2022incontextlearninginductionheads} and retrieval heads~\cite{wu2024retrievalheadmechanisticallyexplains}, with implementation details provided in the Appendix~\ref{app:top_10_induction_retrieval_heads}. The minimal overlap among these and the newly identified RFHs suggests the latter may capture novel patterns not accounted for by known induction or retrieval mechanisms. 

To dive into these reasoning-focused heads, we select one representative head per model and visualize their attention patterns across three sample cases from the MATH-500 dataset, as shown in Figure~\ref{fig:attn_pattern_case_study_MATH500_withR}. 
The positions of key reflection-related tokens, such as \textit{“wait”} and \textit{“alternatively”}, are also marked with vertical lines.

By focusing on the attention region of \textit{Answer} $\rightarrow$ \textit{Reasoning} (highlighted by red boxes), we observe:

\begin{itemize}[leftmargin=5pt]

\item A prominent \textit{attention trajectory} emerges in most cases. This trajectory typically starts at the top-left of the red box and moves diagonally downward (e.g., \textit{``precalculus\_541''}), indicating that as answer generation begins (top of the box), RFHs focus on the beginning of the \textit{Reasoning} segment (left side of the box). As generation progresses, attention shifts accordingly along the reasoning tokens, producing the observed sloped pattern.

\item These trajectories often terminate near the reasoning reflection tokens on the horizontal axis (e.g., \textit{``algebra\_1547''}, \textit{``mid\_algebra\_1898''}, and \textit{``algebra\_893''}). This alignment reflects the model's awareness that such tokens often appear after a solution, signaling a moment for verification or alternative solution. Together with the alignment at the start of the reasoning, this suggests that RFHs closely track the reasoning process and synchronize answer generation with it.

\item Beyond the main trajectory, several parallel attention paths are also observed (e.g., in \textit{``precalculus\_541''}, \textit{``algebra\_2159''}, and \textit{geometry\_538''}). These often originate from and terminate at reflection-related tokens, suggesting they correspond to alternative solutions or verification steps in the \textit{Reasoning} segment.\footnote{In the \textit{``algebra\_2159''} case, the second trajectory starts at a point not marked with a vertical line, due to the model generating a reflection without using explicit keywords like ``wait'' or ``alternatively'' instead saying ``let us double-check...''.} The presence of multiple simultaneous trajectories suggests that RFHs recognize multiple solution paths and attempt to incorporate them during answer generation.

\item Not all attention trajectories are interrupted by reflection tokens. For instance, in \textit{``algebra\_1547''} and \textit{``count \& prob.\_765''}, attention continues despite encountering the first reflection token. This is because reflections can occur in the middle of a solution to re-evaluate specific steps.

\item Lastly, the above observations also apply to the cases with no reflection tokens (e.g., \textit{``algebra\_2476''}) or with excessive amount of reflection tokens (e.g., \textit{``prealgebra\_1865''}). 
\end{itemize}
In summary, our fine-grained case study across selected RFHs reveals that the distilled R1 reasoning models closely follow the \textit{Reasoning} segment during answer generation. They not only align the generation process with the reasoning structure but also exhibit sensitivity to reflective cues, reinforcing their deep integration of the reasoning content.

\subsection{Case Study: Debugging Reasoning Failures with RFHs}

\begin{figure}[t]
  \centering
  \begin{subfigure}[b]{\linewidth}
    \centering
    \includegraphics[width=\linewidth]{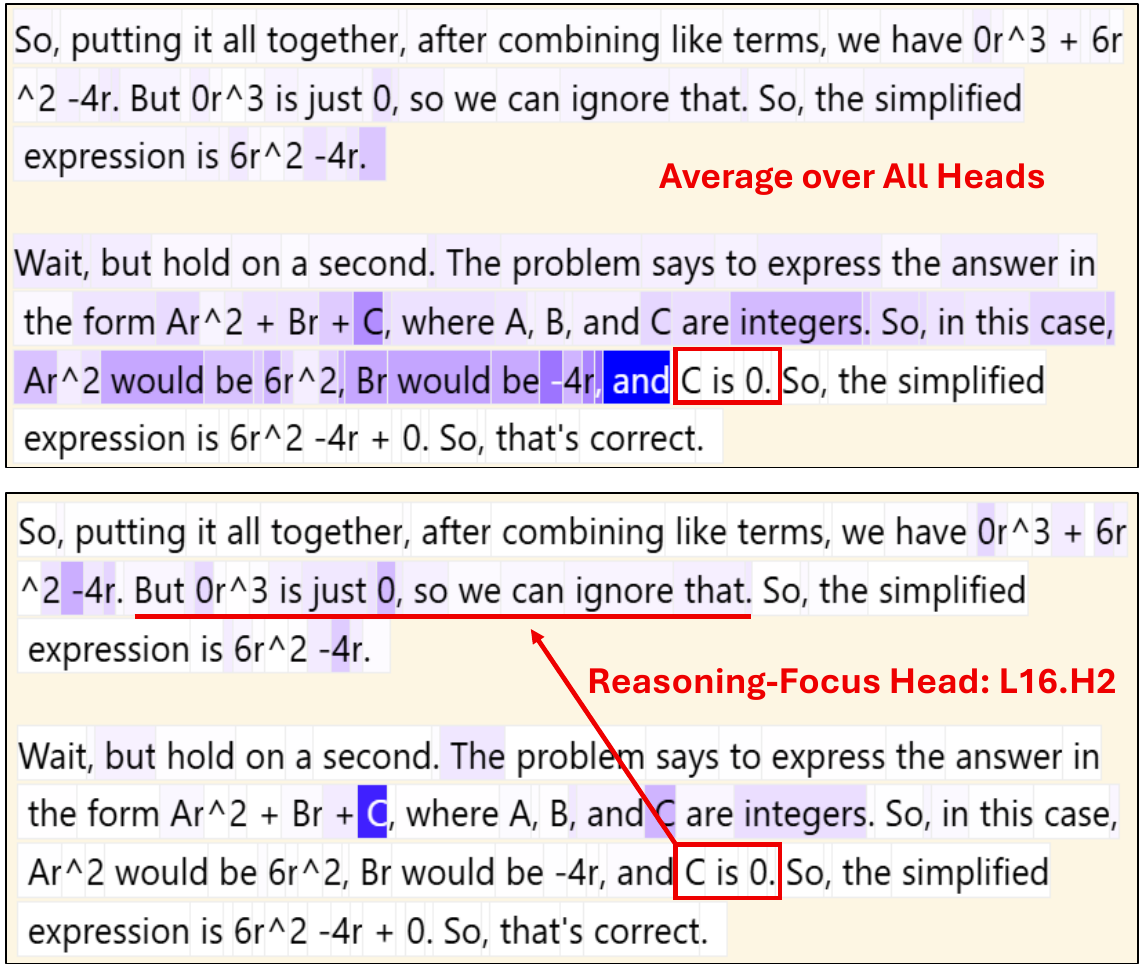}
  \end{subfigure}  
  \caption{Visualization of attention weights for tokens preceding ``\textit{C is 0}'' in the reasoning trace of R1-Qwen-1.5B on \textit{``algebra\_1547''}. The top panel shows attention averaged over all heads, where the focus is largely on nearby tokens. The bottom panel isolates RFH ``L16.H2'', which highlights a strong attention link to the phrase ``\textit{But $0r^3$ is just 0, so we can ignore that}''. Color scales of the top and bottom panels are normalized independently, and if comparing the absolute values of the attention weights, the RFH assigns roughly five times more attention to this phrase than the head-average view.}

  \label{fig:case_study_focused}
\end{figure}

Further investigations suggest that Reasoning-Focus Heads (RFHs) may also provide a useful lens for understanding the model’s reasoning behavior within the \textit{Reasoning} segment, particularly when diagnosing failures in reasoning traces. Figure~\ref{fig:case_study_focused} illustrates an application of RFH ``L16.H2'' in R1-Qwen-1.5B to investigate a failure on the MATH-500 problem \textit{``algebra\_1547''} (see Figure~\ref{fig:attn_pattern_case_study_MATH500_withR} for the overall attention pattern). The problem is: ``\textit{Simplify $4(3r^3+5r-6)-6(2r^3-r^2+4r)$, and express your answer in the form $Ar^2 + Br + C$.}'' The ground-truth answer is $6r^2 - 4r - 24$. The model’s final prediction, however, is $6r^2 - 4r$, having dropped the constant term. Tracing the chain-of-thought reveals an earlier statement, ``\textit{C is 0},'' which propagates to the incorrect final result; yet this statement is not justified by the surrounding verbalized reasoning tokens.

To understand where ``\textit{C is 0}'' originates, we inspect the tokens attended to by the tokens comprising this statement. Figure~\ref{fig:case_study_focused} visualizes attention weights for tokens preceding ``\textit{C is 0}'' under two settings: (i) averaged over all attention heads and (ii) restricted to the identified RFH. In the head-average setting, the majority of the attention mass falls on nearby tokens, providing little insight into the origin of the statement. By contrast, when focusing on the RFH, it becomes apparent that the erroneous conclusion arises because the model strongly attends to the phrase ``\textit{But $0r^3$ is just 0, so we can ignore that.}'' This phrase correctly concerns the vanishing coefficient of the $r^3$ term, but the model conflates it with the constant term, incorrectly inferring ``\textit{C is 0}''. In short, the RFH view makes the source of confusion salient, whereas the head-average view obscures it by focusing local context. 

This observation highlights the potential of RFHs as a practical interpretability tool for model debugging: by isolating reasoning-focus heads, we can more easily identify the origin of reasoning errors and trace them back to specific points in the model’s internal computation. Additional visualizations of the full reasoning-trace attention maps for this example, are provided in Appendix~\ref{app:case_study_failure}.

\section{Mechanistic Intervention: Can Small Reasoning Changes Shift the Answer?}
\label{sec:mechanistic_intervention}

\begin{figure*}[t]
  \centering
  \begin{subfigure}[b]{\textwidth}
    \centering
    \includegraphics[width=\textwidth]{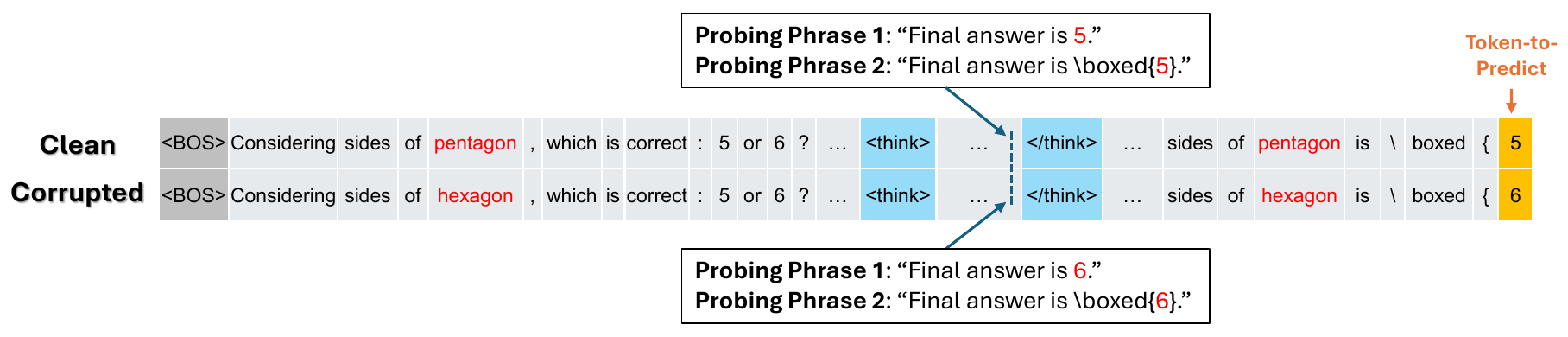}
  \end{subfigure}  
  \caption{Example of aligned clean and corrupted prompts for activation patching. To ensure comparability, the reasoning and answer segments are padded to equal lengths in both clean and corrupted prompts. A probing phrase is inserted at aligned token positions in the reasoning segment using a consistent format. Activation patching is applied to the probing phase, and its effect is measured by the shift in model output at the token-to-predict position.}
  \label{fig:MI_exp_illustration}
\end{figure*}

As strong attention does not guarantee functional dependence, in this section we perform targeted interventions on the reasoning tokens and trace how these modifications affect the model’s output, leveraging mechanistic interpretability tools. Specifically, we use Activation Patching (or Causal Tracing)~\cite{meng2022locating}, which involves running the model on both clean and corrupted versions of a prompt. We intervene in the corrupted run by replacing certain token activations with those from the clean run, then evaluate whether this correction nudges the output closer to the correct answer. By systematically patching activations at various layers for reasoning and answer tokens, we identify which activations are causally important—i.e., those whose restoration substantially increases the likelihood of the correct outcome.

\subsection{Controlled Experiment Settings}

Designing controlled settings for conducting activation patching is not trivial~\cite{heimersheim2024useinterpretactivationpatching}. Here, we introduce a \textbf{Contextual Object Comparison} reasoning task. The format of task query is defined as: “Considering [context], which is [comparator]: [A] or [B]?”, with ``context'' determining how objects A and B are compared. For example, as shown in Figure~\ref{fig:MI_exp_illustration}, a clean query might be “Considering \textit{sides of pentagon}, which is correct: 5 or 6?”, while the corresponding corrupted query is “Considering \textit{sides of hexagon}, which is correct: 5 or 6?”. Since these yield different answers, we can examine how patching specific activations causes the output to flip. To ensure generality, we generate dozens of such query pairs across a variety of domains using the OpenAI \texttt{o1} model. The generation prompt and additional details on data curation are provided in Appendix~\ref{app:prompt_data_curation_COC}.

Using the clean and corrupted query pairs, we instruct reasoning models to generate both reasoning traces and answer tokens, and define the clean and corrupted prompts as the full token sequences comprising the query, instruction, and all generated tokens (mirroring the earlier definition used in the attention analysis). Unlike prior controlled settings (e.g., the Indirect Object Identification task~\cite{2022arXiv221100593W}), our setup presents two specific challenges. First, since the reasoning and answer tokens are generated, the resulting prompt pairs often differ in length. Second, their formatting can diverge, e.g., reasoning traces may or may not conclude with a phrase like \textit{“**Final Answer**:...”}. 

To resolve these issues, we implement a prompt alignment procedure that standardizes the ending phrases in both the reasoning and answer segments across clean and corrupted prompts. Alignment details are provided in Appendix~\ref{app:prompt_alignment}. Figure~\ref{fig:MI_exp_illustration} presents an example of aligned prompts, which now share the same token length and consistent formatting in answer and reasoning segments. We focus on two common concluding phrase formats in reasoning, reflecting frequently observed patterns.


With the aligned prompts, our activation patching experiments are conducted by replacing the activations of reasoning tokens in the clean prompt with those from the corrupted prompt. Since most reasoning tokens in the clean and corrupted prompts differ, we perform activation patching on the reasoning tokens within the probing phase, as well as on a few final answer tokens. This allows us to observe how these answer tokens leverage the newly introduced activation information after patching. To quantify the effect of the intervention, we use the \textit{logit difference}~\cite{heimersheim2024useinterpretactivationpatching}, normalizing it by the raw logit differences of the clean and corrupted prompts (see Appendix~\ref{app:logit_diff} for details). Thus, a value approaching 1 indicates that the answer has flipped, while a value near 0 suggests no impact on the final answer.

\subsection{Intervention Results}

\begin{figure}[t]
  \centering
  \begin{subfigure}[b]{\linewidth}
    \centering
    \includegraphics[width=\linewidth]{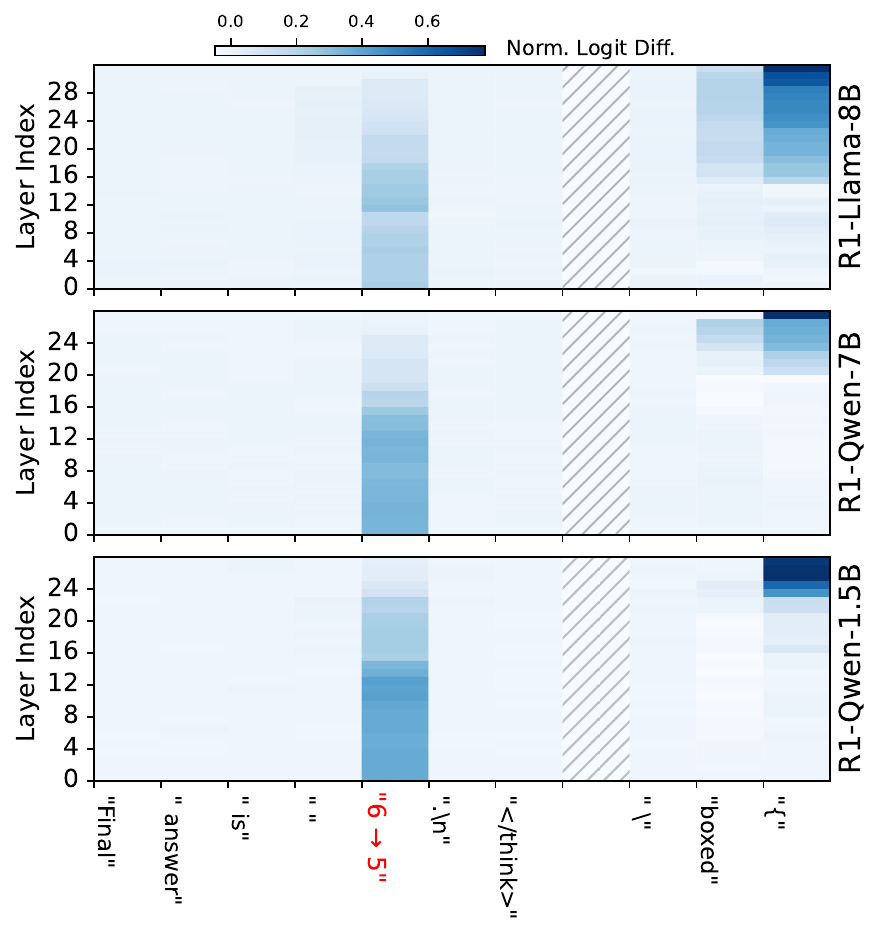}
  \end{subfigure}  
  \caption{Effect of residual stream patching on normalized logit difference for the target token prediction. The heatmaps show the impact of injecting residual streams from the clean prompt into the corrupted prompt at various token positions, measured across model layers. Residual patching is applied at each token in Probing Phrase 1 within the \textit{Reasoning} segment, as well as selected ending tokens in the \textit{Answer} segment.}
  \label{fig:activation_patching_case_heatmap}
\end{figure}

Figure~\ref{fig:activation_patching_case_heatmap} shows the impact of patching residual streams from the clean prompt into the corrupted prompt at each layer for selected reasoning and answer tokens, using the example shown in Figure~\ref{fig:MI_exp_illustration}. We observe that patching the answer-flipping token in reasoning (in red) can increase the normalized logit difference by up to about 0.5. This indicates that such an intervention is highly effective in flipping the answer in the corrupted prompt, providing further evidence that the activations of reasoning tokens can influence the model’s final output.

Furthermore, by comparing the distributions across model layers for the answer-flipping reasoning token and the final two answer tokens (i.e., ``\textit{boxed}'' and ``\{''), we observe that the patching effect on the reasoning token diminishes in later layers, while at similar layers, the patching impact on the answer tokens begins to emerge. This pattern is reminiscent of the Indirect Object Identification task~\cite{2022arXiv221100593W}, where attention heads in these transitional model layers act as information movers—transporting the value of the answer-flipping token and embedding it into the residual stream at the position of the final answer token (e.g., ``\{'').  

\begin{figure}[t]
    \centering
    \includegraphics[width=\linewidth]{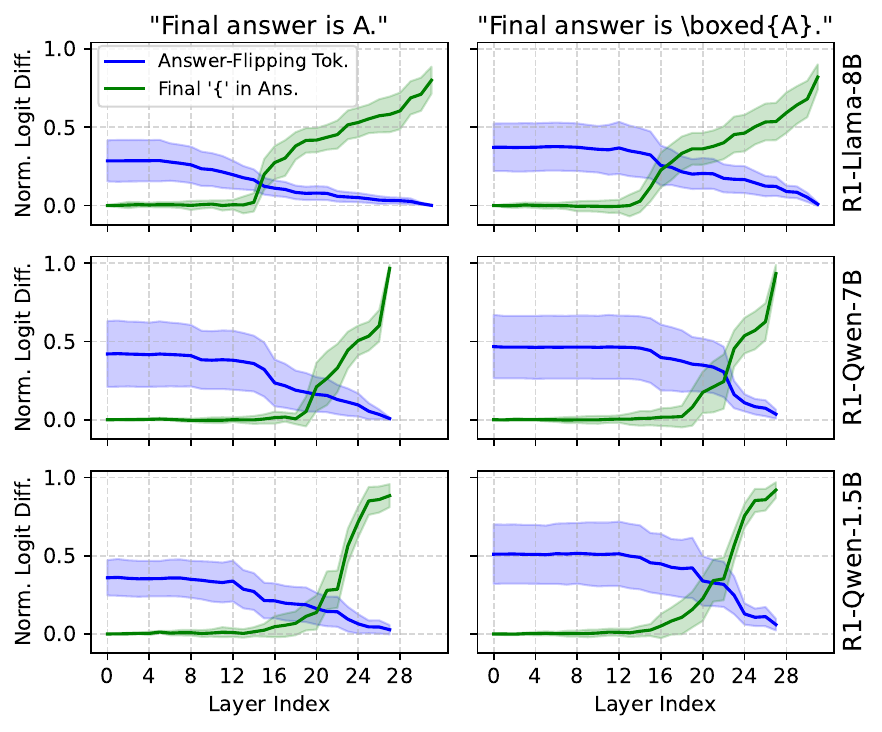}
  \caption{Aggregated results showing how the normalized logit difference evolves across model layers when residual stream patching is applied to specific tokens. Patching is performed at the answer-flipping token in the \textit{Reasoning} segment (blue) and at the preceding answer prediction token (i.e., ``\{'') in the \textit{Answer} segment (green). The left and right columns represent two distinct probing phrases. Shaded regions indicate the distribution across all test samples.}
  \label{fig:patch_resid_pre}
\end{figure}

The above observation generalizes to other test cases within this contextual objection comparison scenario. In Figure~\ref{fig:patch_resid_pre}, we illustrate how the normalized logit difference evolves across model layers when residual stream patching is applied to the answer-flipping reasoning token (blue) and the preceding answer prediction token (green). Two distinct probing phrases are considered, with the shaded bands representing variation across test samples. We observe that these transitional model layers emerge consistently across diverse cases and different probing phrases.

Notably, an astute reader may recognize that these transitional layers approximately correspond to the middle layers identified in our earlier attention analysis, reinforcing the presence of reasoning-focused attention heads. Finally, we find that the patching effect is more pronounced for \textit{Probing Phase 2}, which includes the ``\textbackslash boxed\{'' tag. This heightened effect may arise from the recurrence of the same tag in the answer prediction token, forming a pattern of \texttt{``...ab...a$\rightarrow$b''}, where \texttt{a} and \texttt{b} correspond to ``\textbackslash boxed\{'' and the final answer token, respectively. This recurrence potentially activates the induction head~\cite{olsson2022incontextlearninginductionheads}, thereby facilitating additional information propagation.

\section{Conclusion}

We presented a multi-faceted investigation into how reasoning traces influence answer generation in large reasoning models, focusing on distilled variants of DeepSeek R1. Our study combined empirical evaluation, attention analysis, and mechanistic intervention to assess whether and how models leverage reasoning tokens during inference. We find that explicit reasoning improves answer quality across diverse domains. From the attention analysis, answer tokens consistently attend to reasoning segments; moreover, within this analysis, \emph{Reasoning-Focus Heads} (RFHs) emerge as heads that track the reasoning process (including self-reflection) and make failure modes interpretable than head-averaged views. Finally, mechanistic interventions show that small changes to reasoning activations can flip final outputs. 

Together, these results provide converging evidence for a functional dependence between reasoning and answers, shedding light on the internal dynamics of LRMs and informing efforts to improve faithfulness, controllability, and monitoring.

\section*{Limitations}

\noindent \textbf{Model Scope.} Our analyses focus on three distilled variants of DeepSeek R1 due to their accessibility and tractable size. Although these models exhibit consistent trends across tasks, our findings may not generalize to other LRMs such as the full R1 model or OpenAI o1. Future work should examine whether similar reasoning-to-answer dependencies exist across a broader range of LRMs.

\noindent \textbf{Dataset Coverage and Scale.} Although our empirical evaluation spans both math (MATH-500) and open-domain tasks (WildBench), the number of test samples per model is moderate due to computational constraints. While qualitative trends are consistent, more extensive benchmarking would help strengthen statistical confidence in observed effects, particularly for open-ended domains.

\noindent \textbf{Controlled Interventions.} Our mechanistic interventions rely on clean-corrupted prompt pairs in a controlled, synthetic reasoning format. While this design allows for precise attribution of answer changes, it may not capture the full range of naturalistic perturbations or adversarial reasoning errors encountered in real-world deployments. Generalizing the intervention results to unconstrained inputs remains an open challenge.

\noindent \textbf{Depth of Mechanistic Analysis.} Our mechanistic interpretability work operates at the residual stream level using activation patching. While this reveals causal influence from reasoning tokens to answers, we do not perform in-depth circuit-level tracing or path attribution to identify the precise internal substructures responsible for this information flow. Future work could build on our findings to discover and characterize the underlying circuits that mediate reasoning integration.

\section*{Ethics Statement}

This work analyzes publicly available models and datasets (DeepSeek R1 variants, MATH-500, WildBench) for research purposes only. No personally identifiable or sensitive data is used. Our goal is to improve transparency and understanding of model reasoning, not to deploy models in real-world applications. While we employ interpretability tools like activation patching, we caution that such methods must be used responsibly. We release code at \href{https://aka.ms/R2A-code}{this URL} to support reproducibility.

\section*{Acknowledgments}
We thank Fangkai Yang, Xiaoting Qin, Zhitao Hou, and Yi Ren for their insightful discussions and feedback. We are also grateful to Bo Qiao for assistance in setting up the experimental environment. We further thank the anonymous reviewers for their careful reading and helpful suggestions, which greatly improved the quality of this work. Finally, we gratefully acknowledge the developers of the open-source \texttt{transformer\_lens} package~\cite{nanda2022transformerlens}, which provided a crucial foundation for our analysis.

\bibliography{main}

\appendix

\section{Implementation Details for Empirical Evaluation}
\label{app:quality_impact}

\begin{table*}[h!]
\centering
\small
\renewcommand{\arraystretch}{1.2} 
\begin{tabular}{c|c|c|c|c|c}
\hline
\textbf{Setting} & \textbf{Num} & \textbf{\makecell[c]{+Finished\\(once / twice)}} & \textbf{\makecell[c]{+ThinkFormat\\(once / twice)}} & \textbf{\makecell[c]{+AnswerFormat\\(once / twice)}} & \textbf{\makecell[c]{+SameId\\(once)}} \\
\hline
R1-Llama-8B-withR     & 500 & 412 / 448 & 379 / 433 & 373 / 429 & 355 \\
R1-Llama-8B-withoutR  & 500 & 490 / 498 & 489 / 498 & 465 / 490 & 355 \\
R1-Qwen-7B-withR      & 500 & 470 / 483 & 470 / 483 & 466 / 479 & 457 \\
R1-Qwen-7B-withoutR   & 500 & 496 / 499 & 490 / 497 & 483 / 492 & 457 \\
R1-Qwen-1.5B-withR    & 500 & 421 / 454 & 421 / 454 & 415 / 448 & 400 \\
R1-Qwen-1.5B-withoutR & 500 & 483 / 494 & 474 / 493 & 459 / 484 & 400 \\
R1-Full-withR      & 500 & 500 / - & 499 / - & 493 / - & 490 \\
R1-Full-withoutR   & 500 & 500 / - & 499 / - & 491 / - & 490 \\
\hline
\end{tabular}
\caption{Step-by-step sample counts after applying the data filtering process on the MATH-500 dataset. ``+Finished'' indicates the number of samples that complete within the token limit, ``+ThinkFormat'' enforces the presence of the \texttt{<think>} tag, ``+AnswerFormat'' ensures that the final answer is enclosed within ``\texttt{\textbackslash boxed\{\}}'', and ``+SameId'' aligns the sample set across reasoning and non-reasoning conditions. Results are reported after two independent runs (``once'' and ``twice'') to illustrate the effect of model randomness on format conformance.}
\label{tb:data_filtering}
\end{table*}

To generate the evaluation results presented in Section~\ref{sec:empirical_evaluation}, we first deploy the three distilled R1 models locally, while utilizing the Azure-hosted API endpoint for the full DeepSeek R1 model. Following the recommendations from DeepSeek~\cite{deepseekai2025deepseekr1incentivizingreasoningcapability}, we append the instruction \textit{``Please reason step by step, and put your final answer within \textbackslash boxed\{\}.''} to all math-related queries (i.e., the MATH-500 dataset) and set the temperature to 0.6 for all R1 variants. Zero-shot prompting is employed for both the MATH-500 and WildBench datasets. Due to resource limitations, we cap the maximum output length at 10k tokens for the distilled models and 32k tokens for the full R1 model.

After collecting the model outputs, we perform post-processing by discarding responses that are incomplete or do not conform to the required format. For example, Table~\ref{tb:data_filtering} presents the detailed sample counts after each filtering step when evaluating on the MATH-500 dataset. Specifically, ``+Finished'' indicates the number of samples that complete within the token limit, ``+ThinkFormat'' refers to the additional filtering step that ensures the response contains the correct \texttt{<think>} tag, ``+AnswerFormat'' verifies that the final answer is wrapped in ``\texttt{\textbackslash boxed\{\}}'', and ``+SameId'' ensures that the same set of samples is used for both the reasoning and non-reasoning settings. 

To examine how model randomness affects format conformance, we perform two runs with identical settings for the three distilled R1 models on the MATH-500 dataset and report the sample counts after the first and second attempts as ``once'' and ``twice'' in Table~\ref{tb:data_filtering}. As shown, a second attempt successfully recovers additional samples, particularly for Llama-8B with reasoning, where the number of valid samples increases by approximately $10\%$. 

The same data filtering procedure is applied to the WildBench dataset, with the exception that math-related queries are excluded due to the lack of precise ground-truth answers. Furthermore, to mitigate variance caused by limited data in certain sub-task types, we aggregate the WildBench task domains into four broader categories~\cite{Lin2024WildBenchBL} using the following mapping:

{\footnotesize
\begin{itemize}[leftmargin=5pt]
    \item ``Coding \& Analysis'': ``Coding \& Debugging'', ``Data Analysis''
    \item ``Creative Tasks'': ``Brainstorming'', ``Creative Writing'', ``Editing'', ``Role Playing''
    \item ``Information Seeking'': ``Advice Seeking'', ``Information Seeking''
    \item ``Reasoning \& Planning'': ``Reasoning'', ``Planning''
\end{itemize}
}
\noindent The resulting number of valid samples (after the first model run) used in our evaluation is shown in Figure~\ref{fig:quality_impact}.

For model output evaluation, we follow the methodology provided with the MATH-500 and WildBench datasets. Specifically, MATH-500 is evaluated using rule-based answer extraction and matching, whereas WildBench relies on an LLM judge (``GPT-4o-20240513'' in our case). Statistical significance of the difference between the with- and without-reasoning conditions is assessed using a paired t-test.

Finally, to evaluate the effect of temperature, we experimented with greedy decoding (i.e., temperature = 0) and observed results that were largely consistent with those in Figure~\ref{fig:quality_impact}. For example, Table~\ref{tb:greedy_results} reports the accuracy results on the MATH-500 dataset under this decoding strategy.

\begin{table}[h!]
\centering
\small
\begin{tabular}{lcc}
\hline
\textbf{Model Type} & \textbf{\makecell[c]{Accuracy\\(withR / withoutR)}} & \textbf{p-value} \\
\hline
R1-Llama-8B  & 94.1\% / 77.5\%  & $p < 0.001$ \\
R1-Qwen-7B   & 95.4\% / 84.8\%  & $p < 0.001$ \\
R1-Qwen-1.5B & 95.6\% / 86.8\%  & $p < 0.001$ \\
\hline
\end{tabular}
\caption{Greedy decoding results for MATH-500.}
\label{tb:greedy_results}
\end{table}

\section{Attention Analysis for WildBench}
\label{app:attn_analysis_wildbench}

\begin{figure}[t]
  \centering
    \includegraphics[width=\linewidth]{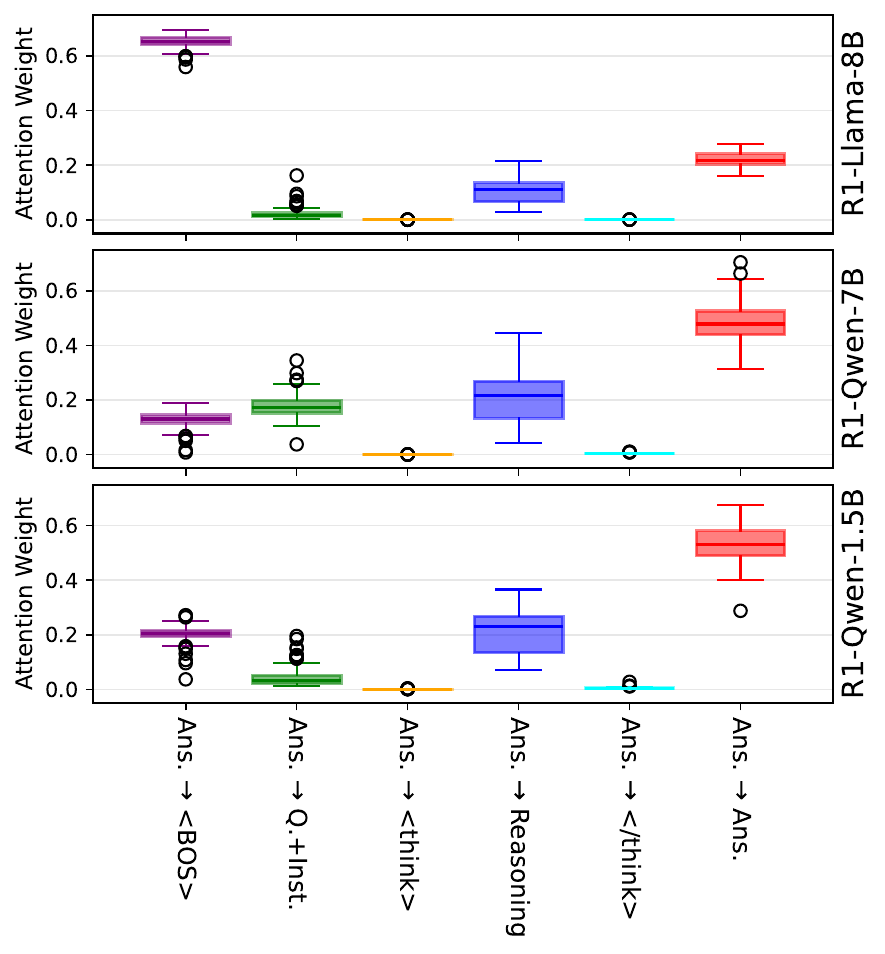}    
  \caption{Decomposition of average attention weights from answer tokens to different prompt segments across three models (R1-Llama-8B, R1-Qwen-7B, R1-Qwen-1.5B) using the WildBench dataset.}
  \label{fig:answer_attn_decompose_WildBench}
\end{figure}

\begin{figure*}[t]
    \centering
    \includegraphics[width=\linewidth]{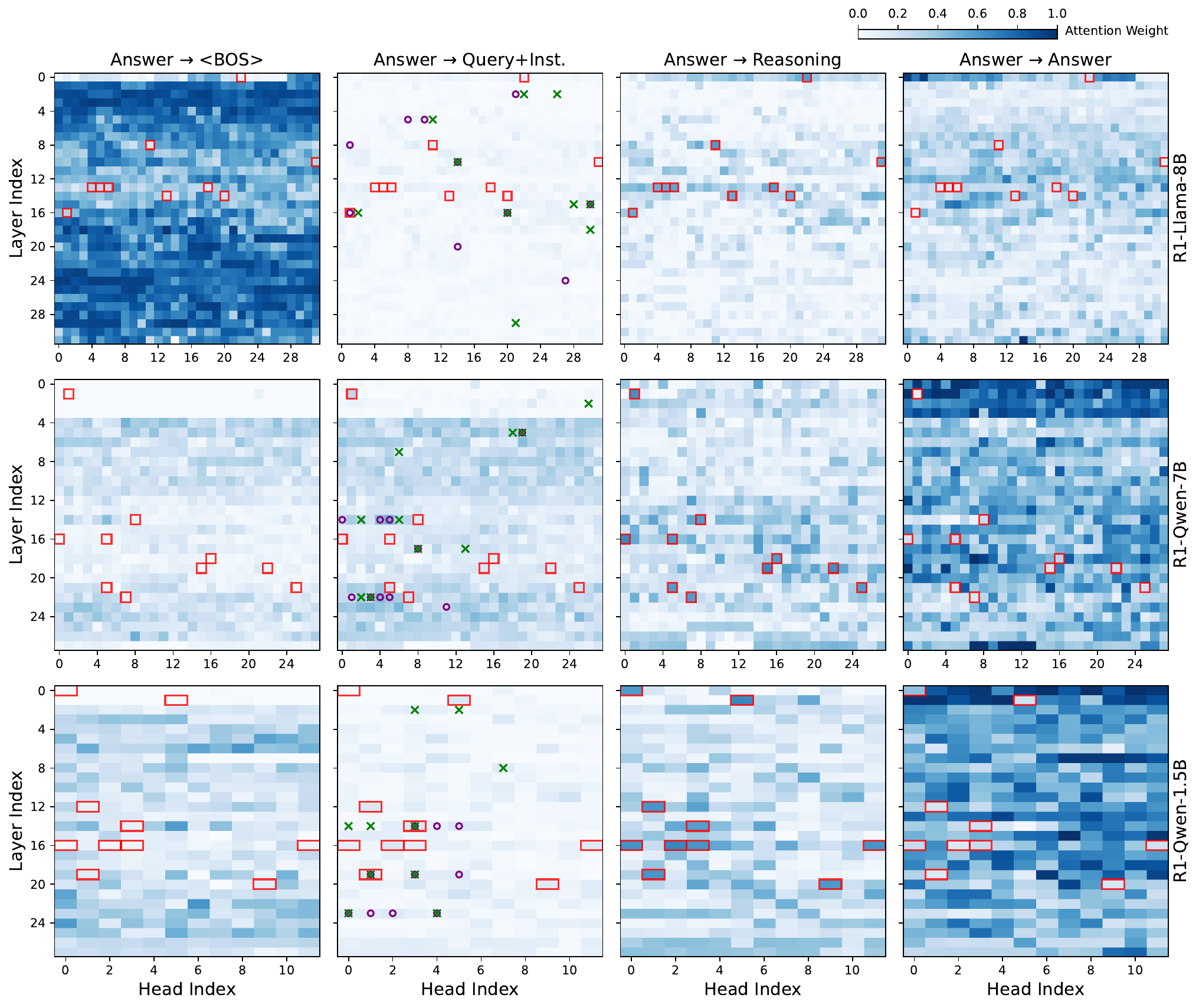}
  \caption{Decomposition of answer attention to different prompt segments across layers and heads for three models (R1-Llama-8B, R1-Qwen-7B, and R1-Qwen-1.5B) using the WildBench dataset. Heatmaps show the attention weights from the \textit{Answer} segment to four prompt segments: \texttt{<BOS>}, \textit{Query+Instruction}, \textit{Reasoning}, and \textit{Answer}. Red boxes highlight the top 10 attention heads with the highest weights for \textit{Answer} $\rightarrow$ \textit{Reasoning}. Additionally, in the second column the top 10 retrieval and induction heads are annotated with ``{\color{purple}{$\boldsymbol{\circ}$}}'' and ``{\color{darkgreen}{$\boldsymbol{\times}$}}'', respectively.}
  \label{fig:attn_map_WildBench}
\end{figure*}

Here we present attention analysis results on the WildBench dataset, complementing the corresponding results on the \textsc{MATH-500} dataset discussed in the main text. Segment-level attention patterns are shown in Figure~\ref{fig:answer_attn_decompose_WildBench}, while detailed attention head-level results are provided in Figure~\ref{fig:attn_map_WildBench}. The overall trends are consistent with those observed in \textsc{MATH-500}, with two notable differences. First, the aggregated attention from answer to reasoning segments (blue box-plots) exhibits greater variation in WildBench, likely due to its broader domain diversity and corresponding variability in reasoning patterns. Second, the top ten attention heads focusing on reasoning (red boxes in Figure~\ref{fig:attn_map_WildBench}) differ slightly from those in \textsc{MATH-500}. Eight of these heads overlap with the top heads in \textsc{MATH-500}, while the remaining two, though not in the top ten for \textsc{MATH-500}, still exhibit relatively high answer-to-reasoning attention in that dataset.

\section{Attention Patterns for Selected Attention Heads in Early Model Layers}
\label{app:early_layer_attn_head}

\begin{figure}[t]
    \centering
    \includegraphics[width=\linewidth]{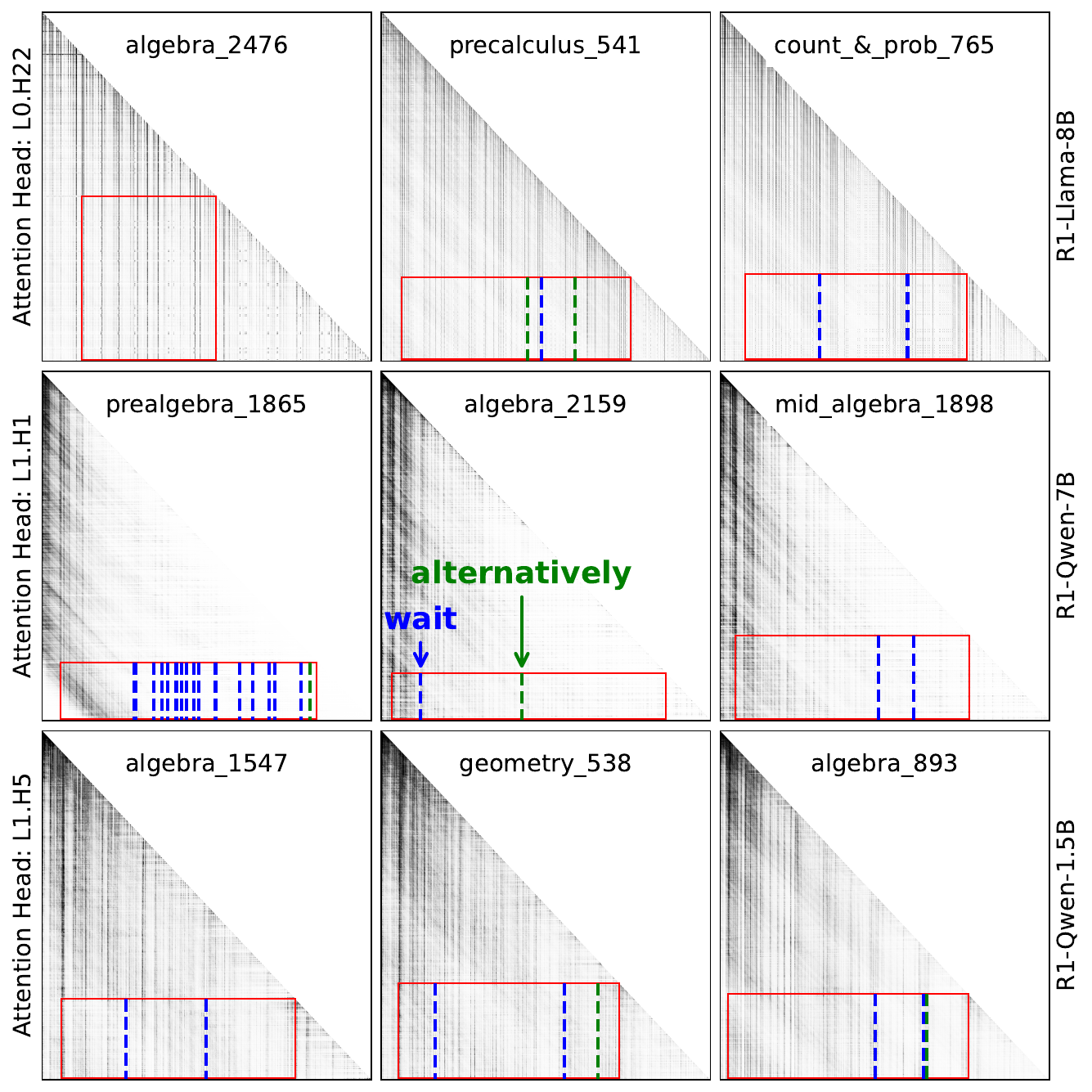}
    \caption{Attention patterns from selected attention heads in the early layers of the three distilled R1 models on sample cases from the MATH-500 dataset. The horizontal and vertical axes represent key and query token indices, respectively. Labels on the left y-axis denote the selected reasoning-focused heads (e.g., \textit{L0.H22''} refers to head 22 in Layer 0). The \textit{Answer} $\rightarrow$ \textit{Reasoning} region is highlighted with red boxes. Vertical lines indicate the positions of the tokens ``\textcolor{darkblue}{wait}'' and ``\textcolor{darkgreen}{alternatively}''.}
  \label{fig:attn_pattern_case_study_early_layer_MATH500_withR}
\end{figure}

In Figure~\ref{fig:attn_pattern_case_study_early_layer_MATH500_withR}, we present attention patterns from three selected attention heads in the early layers of the three distilled R1 models. The sample cases are identical to those shown in Figure~\ref{fig:attn_pattern_case_study_MATH500_withR}. A comparison between Figure~\ref{fig:attn_pattern_case_study_early_layer_MATH500_withR} and Figure~\ref{fig:attn_pattern_case_study_MATH500_withR} reveals that although these early-layer heads also assign high attention from answer tokens to reasoning tokens, their attention maps exhibit no discernible structure, instead showing a near-uniform distribution. This suggests that such patterns are not indicative of genuine reasoning focus, but rather reflect artifacts related to the token length.

\section{Implementation Details for Obtaining Induction and Retrieval Heads}
\label{app:top_10_induction_retrieval_heads}

The induction heads for the three distilled R1 models are identified using the \texttt{detect\_head()} function from the \texttt{transformer\_lens} Python library~\cite{nanda2022transformerlens}. Sample prompts include \textit{``one two three one two three one two three''}, \textit{``1 2 3 4 5 1 2 3 4 1 2 3 1 2 3 4 5 6 7''}, and \textit{``green ideas sleep furiously; green ideas don't sleep furiously''}. Default settings are used, and alternative configurations were also tested, yielding negligible differences in output.

For the retrieval heads, we adopt the open-source implementation provided with~\cite{wu2024retrievalheadmechanisticallyexplains}, with minor modifications to support the three distilled R1 models. We use the default configuration, setting the detection length to 5000 (i.e., \texttt{--e 5000}) to accommodate our GPU constraints.

\section{Additional Case Details for Debugging Reasoning Failures with RFHs}
\label{app:case_study_failure}

\begin{figure*}[t]
    \centering
    \includegraphics[width=\linewidth]{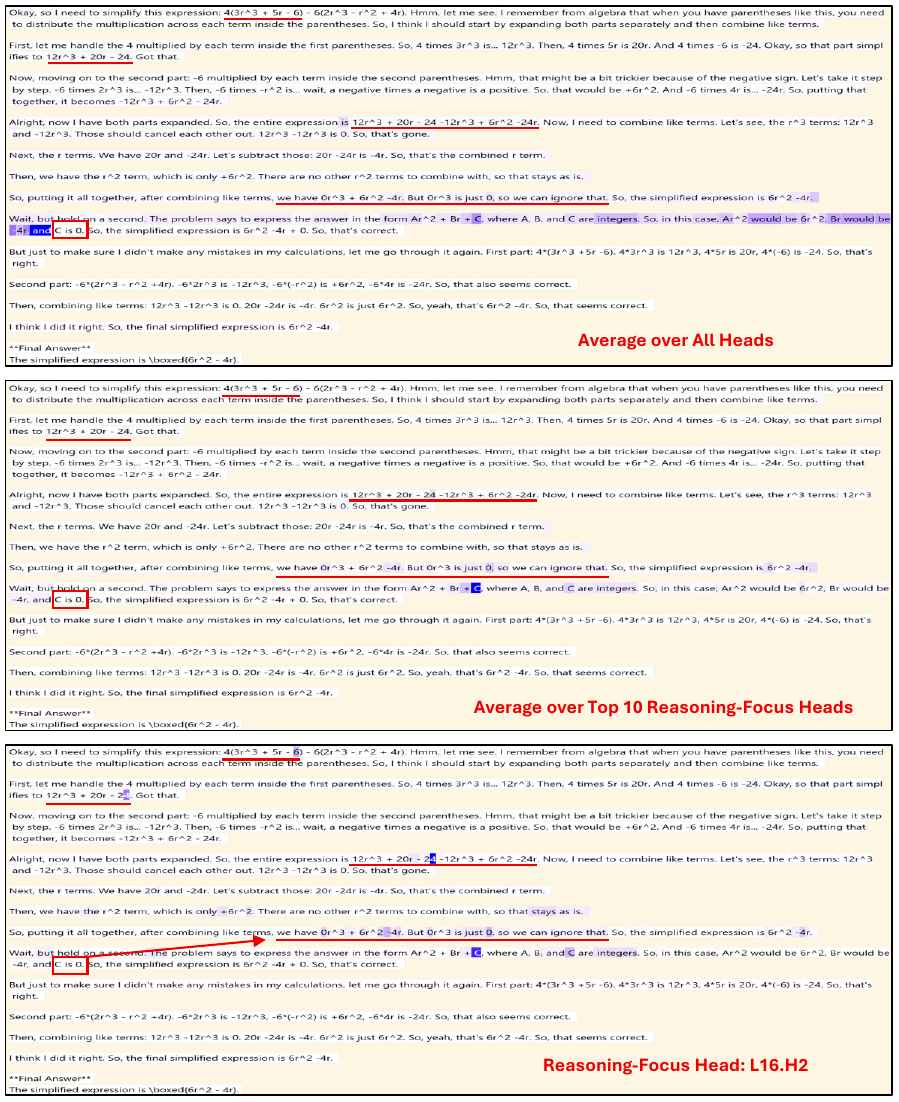}
  \caption{Visualization of attention weights across the entire \textit{Reasoning} segment for three settings: (top) average over all heads, (middle) average over the top 10 Reasoning-Focus Heads (RFHs) identified in R1-Qwen-1.5B, and (bottom) the specific RFH ``L16.H2''. While the head-average view primarily focuses on local tokens, the RFH views highlight strong attention to the phrase ``\textit{But $0r^3$ is just 0, so we can ignore that.}'', revealing that the model confuses the vanishing cubic term with the constant term when generating ``\textit{C is 0}''. Interestingly, the RFH view also shows strong attention to the correct constant term coefficient ``$-24$,'' suggesting that the necessary information is present but may be misprocessed by the downstream feedforward module, ultimately leading to the incorrect conclusion that $C=0$.}

  \label{fig:case_study_full}
\end{figure*}

Figure~\ref{fig:case_study_focused} in the main text visualizes attention weights for a limited set of tokens preceding the error phrase ``\textit{C is 0}.'' Here, we extend this analysis by visualizing attention over \emph{all} reasoning tokens in the \textit{Reasoning} segment (Figure~\ref{fig:case_study_full}) and by introducing an additional setting that averages over the top 10 Reasoning-Focus Heads (RFHs) identified in R1-Qwen-1.5B (Figure~\ref{fig:attn_map_MATH500}). In these visualizations, the top, middle, and bottom panels correspond to (i) the all-head-average view, (ii) the average over the top-10 RFHs, and (iii) the view for the specific RFH ``L16.H2,'' respectively. All visualizations in Figures~\ref{fig:case_study_focused} and~\ref{fig:case_study_full} are obtained by utilizing the \texttt{circuitsvis} package~\cite{cooney2023circuitsvis}.

Comparing these views highlights the value of RFHs for interpretability. While the head-average view primarily attends to local context, the RFH views reveal that the model strongly links the error phrase ``\textit{C is 0}'' to the statement ``\textit{But $0r^3$ is just 0, so we can ignore that.}'', indicating that the model confuses the vanishing cubic coefficient with the constant term. Interestingly, the RFH view also attends to the correct constant term coefficient ``$-24$'', suggesting that the relevant information is available but is ultimately misprocessed. This observation hints that the subsequent feedforward module in the transformer block may play a critical role in propagating or distorting the attended information, potentially leading to the incorrect conclusion that $C=0$. Exploring how these downstream modules transform attention-derived signals presents an important direction for future work.

\section{Data Curation for Contextual Object Comparison Scenario}
\label{app:prompt_data_curation_COC}

Table~\ref{tab:prompt_data_curation_COC} presents the prompt used to generate sample query pairs for the Contextual Object Comparison scenario described in the main text. The prompt enforces two main constraints: i) candidate answers `A' and `B' must be single-token words or numbers. This simplifies evaluation by focusing on the prediction of a single token and facilitates the computation of logit differences between the two candidates. ii) the query domains are designed to be diverse. Approximately 30\% of query pairs involve numerical answers, 10\% are binary choice queries, and the remainder involve arbitrary one-token words. This distribution mitigates bias arising from varying types of candidate objects.

We begin by generating 200 query pairs using OpenAI o1. To ensure that the candidate answers consist of a single token, we apply the tokenizers of the three R1 distilled models and retain only those pairs that meet this criterion, resulting in 103 valid query pairs. Each model then generates a response for these pairs, with the appended instruction \textit{``Please reason step by step (but not overthinking), and put your final answer within \textbackslash boxed{}''}. We also constrain the maximum output length to 3,000 tokens to fit within our computational limits. After generation, we parse the responses and discard those that exceed the token budget. Ultimately, we obtain 59 query pairs for R1-Llama-8B, 67 for R1-Qwen-7B, and 22 for R1-Qwen-1.5B. The smaller number for R1-Qwen-1.5B arises from its longer reasoning outputs, which often exceed the token limit. Increasing the budget to 5,000 tokens did not resolve this issue.

\begin{table*}[htbp]
    \centering
    \small
    \begin{tabular}{p{15cm}}
    \toprule
    You are an experienced query data generator. I want you to generate 200 query pairs.\\
    \\
    \# On the query format: \\
    Each query pair (`Query-a' and `Query-b') should have the following consistent format: \\    
      - Query-a: Considering \{condition\_a\}, which is \{comparator\}: \{A\} or \{B\}?\\
      - Query-b: Considering \{condition\_b\}, which is \{comparator\}: \{A\} or \{B\}?\\
    \\
    \# Instructions for query pair generation:\\    
    - \{condition\_a\} and \{condition\_b\} are two distinct conditions so that the answer for `Query-a' should always be `A' while the answer for `Query-b' is always `B'.\\
    - \{condition\_a\} and \{condition\_b\} should have the same token length. By token length, I mean the number of tokens if passing \{condition\_a/b\} to an LLM tokenizer.\\
    - \{comparator\} can be any comparison-related word, e.g., ``greater''.\\
    - \{comparator\} should be the same for both queries in the query pair.\\
    - Candidate answers of \{A\} and \{B\} need to be one-token word or number.\\
    - If candidate answers of \{A\} and \{B\} are words, they MUST be simple short 1-token words. For example, 'cheetah' is not allowed for the candidate answers, as it is split into 3 tokens in most LLMs.\\
    - If candidate answers of \{A\} and \{B\} are numbers, they MUST be single-digit numbers (i.e., < 10).\\
    - For diversity, I want the total query pairs to cover diverse domains. Avoid to simply replace the candidate answers without major changes on the corresponding conditions.\\
    - About 30\% of query pairs should have the candidate answers of \{A\} and \{B\} to be single-digit numbers (i.e., < 10).\\
    - About 10\% of query pairs should be binary choice queries, i.e., \{A\} and \{B\} are `yes' and `no'.\\
    \\
    \# On the output format:\\
    - Output should be organized as a table form.\\
    - The output table should have columns of ``query\_a", ``answer\_a", ``query\_b", ``answer\_b", ``domain", ``explanation".\\
    - ``domain'' field provides the query domain.\\
    - ``explanation'' provides a brief explanation of why `query\_a/b' should have distinct answers of `answer\_a' and `answer\_b'.  \\  
    \bottomrule
    \end{tabular}
    \caption{Prompt used for generating query pairs in the Contextual Object Comparison scenario.}
    \label{tab:prompt_data_curation_COC}
\end{table*}

\section{Alignment Procedure for Clean and Corrupted Prompts}
\label{app:prompt_alignment}

To align the clean and corrupted prompts, we first standardize the concluding phrases in the \textit{Answer} and \textit{Reasoning} segments. This involves removing existing variations and replacing them with a consistent format. For the \textit{Answer} segment, we adopt the consistent concluding phrase \textit{``Thus, the \{comparator\} \{condition\} is \textbackslash boxed\{''}, where the first and second placeholders are replaced with the corresponding comparator and condition for each sample, respectively. For the \textit{Reasoning} segment, we use two different concluding phrases, illustrated in Figure~\ref{fig:MI_exp_illustration}, which also serve as probing phrases in the activation patching experiments.

After standardization, we equalize the token lengths of clean and corrupted prompts through \textit{segment-wise} token padding. Left padding is applied within each segment using the model's default padding token. This alignment may alter the model's original output, potentially changing the predicted answer token. To control for this, we discard samples where the absolute logit difference exceeds 5 after alignment.

\section{Definition of Normalized Logit Difference}
\label{app:logit_diff}

We first define the logit difference $LD$ for each prompt as follows,
\begin{equation}
    LD = logit(A) - logit(B),
\end{equation}
where $logit(A)$ is the logit value for the candidate answer token `A'. With this, we then define the Normalized Logit Difference ($NLD$) after patching as:
\begin{equation}
    NLD(LD) = \frac{LD - LD_{corrupted}}{LD_{clean} - LD_{corrupted}},
\end{equation}
where $LD_{clean}$ and $LD_{corrupted}$ are the logit difference for the original clean and corrupted prompts before patching, respectively, and $LD$ is the logit difference of the corrupted prompt after patching.

\end{document}